\documentclass{article}

\usepackage{ijcai16}

\usepackage{times}
\usepackage{url}

\usepackage{amsmath,amssymb}
\usepackage{booktabs}

\usepackage{tikz}
\usetikzlibrary{shapes}

\newtheorem{definition}{Definition}

\newcommand{\Omit}[1]{}

\newcommand{\Qsub}[2]{\mathbb{P}_{#1}(#2)}
\newcommand{\Qh}[1]{\Qsub{h}{#1}}

\newcommand{\simpleP}[1]{\mathbb{P}(#1)}
\renewcommand{\P}[2]{\mathbb{P}(#1|#2)}

\newcommand{\tup}[1]{\langle #1 \rangle}
\newcommand{\R}{\mathcal{R}}
\newcommand{\F}{\mathcal{F}}
\newcommand{\Alert}[1]{\textcolor{red}{\small #1}}

\newcommand{\ACm}{$\textrm{AC}_m$}
\newcommand{\ACzero}{$\textrm{AC}_0$}
\newcommand{\ACone}{$\textrm{AC}_1$}

\newcommand{\citeay}[1]{\citeauthor{#1} [\citeyear{#1}]}

\allowdisplaybreaks

\begin{document}

\title{Factored Probabilistic Belief Tracking}
\author{Blai Bonet\\
        Universidad Sim\'on Bol\'{\i}var\\
        Caracas, Venezuela \\
        {\normalsize\url{bonet@ldc.usb.ve}}
\And
        Hector Geffner \\
%       Department of Information and Commnication Technologies (DTIC) \\
        ICREA \&  Universitat Pompeu Fabra \\
        Barcelona, SPAIN \\
        {\normalsize\url{hector.geffner@upf.edu}}
}

\pdfinfo{
/Title (Factored Probabilistic Belief Tracking)
/Author (Blai Bonet, Hector Geffner)
}

\maketitle

\begin{abstract}
The  problem of  belief tracking in the presence of stochastic actions and observations is pervasive and yet computationally intractable. 
% The  most common approximations are \emph{particle filtering} methods based on state samples, and \emph{heuristic decomposition} methods that  project globals beliefs  
% into local factors. The former  however may require too many samples, while the latter may result  into poor approximations. 
In this work we show  however that  \emph{probabilistic beliefs can be maintained in factored form exactly and efficiently} across a number of 
causally closed beams,  when the state variables  that appear  in more than one beam obey a form of \emph{backward determinism}.  
Since computing marginals from the factors is still  computationally intractable in general, and variables appearing in several beams are not always backward-deterministic, 
the basic formulation is extended with two approximations:  forms of \emph{belief propagation}  for  computing marginals from  factors, and \emph{sampling} of  
non-backward-deterministic variables for making such variables backward-deterministic \emph{given their sampled history.} Unlike, Rao-Blackwellized
particle-filtering, the sampling  is  not used for  making inference  \emph{tractable} but  for making the  factorization \emph{sound}.  
The resulting algorithm  involves  sampling \emph{and} belief propagation or just one of them as determined by the structure of the model.
% and it is  tested on  domains that illustrate these various possibilities.
%  including simultaneous localization and mapping problems
\end{abstract}

\section{Introduction}

Keeping track of beliefs when actions and sensors are probabilistic is  crucial and yet computationally intractable, 
with exact algorithms running 
in time that is exponential in the number of state variables in the worst case. Current approaches rely on 
samples for approximating probabilistic beliefs by sets of particles  \cite{kanazawa:pf,thrun:pf}  or  decompositions where
joint beliefs are approximated by products of smaller local beliefs \cite{bk}.   
Particle filtering methods however  may require too many particles even in the Rao-Blackwellized (RB) version
\cite{murphy:1999,rb}, while decomposition approaches may  result in  poor approximations. 

In this  work we take a different approach and show  that  probabilistic beliefs can be maintained in 
\emph{factored form   exactly and efficiently across a number causally closed beams} when the state variables  that appear  
in more than one beam obey a form of \emph{backward determinism}  by which the value of a   variable at time $t$ 
is determined by the  value of the variable at  time $t+1$,  the history,  and the prior beliefs. Since computing marginals from a  
factorized representation is computationally hard in general, and variables appearing in several beams are not always backward-deterministic, 
the basic formulation is extended  through two  approximations:  forms of  \emph{belief propagation}  for  computing marginals 
from the  factorized representation  \cite{pearl:book}, and \emph{sampling} of  non-backward-deterministic variables
for making them backward-deterministic \emph{given their sampled history.} This last part is 
similar to Rao-Blackwellized particle-filtering with one crucial difference:  sampling  is  not introduced  for  making
inference  \emph{tractable} but  for making the factorization \emph{sound}.
Like the method of \citeay{bk},
the algorithm maintains  global beliefs in terms of smaller local  factors  but the scope of these  local factors is determined 
by the structure of the model and  variables usually appear in many factors. %\Alert{(PBT doesn't do projections)} revised

We call the general algorithm \emph{probabilistic beam tracking} (PBT) as it is the probabilistic version of the 
\emph{beam tracking} scheme of \citeay{bonet:jair2014} that deals with beliefs represented as \emph{sets} of states
rather than  probability distributions. When the size of the beams and the number of particles are  bounded, 
PBT runs in \emph{polynomial time}. In addition, when  the structure of the beams results
in factored representations that are \emph{acyclic} and where factors share variables that are  backward-deterministic only,
the algorithm is \emph{exact}. 

%%% no importante que lo de abajo es una reformulacion de murphy; ya luego lo hablamos, aqui lo dejamos asi
As an illustration, the 1-line SLAM problem of \citeay{murphy:1999} that involves an agent that moves
along a line  sensing  the color of  each cell $i$, results in  factors $B_i$ of size two, where one variable 
represents  the agent location,  and the other,  the color of cell $i$.  PBT reduces then  to RB particle-filtering, 
as the variable that appears in more than one beam (the agent location)  is not backward deterministic and must be sampled,  
leaving each beam with a single, unique variable. On the other hand, if the  problem is modified so that the color sensed in a cell depends on the color of 
the two surrounding cells, PBT would  still sample the agent location variable, but in addition would keep
track of  factors of size three representing the color of each cell and the color of the two surrounding cells.
Inference over these factors can be done exactly by the jointree algorithm, as the treewidth of the factor
graph is bounded and small \cite{darwiche:book}, or  approximately but more efficiently in general, by belief propagation.
In all cases, the formulation determines the scope of the factors (the beams) and the set of variables that need to be sampled
from the structure of the model. 

We begin by discussing the background and related work, and then present the model,
the structure, and the beams that follow from them.
Next, we derive the equations for factorized belief tracking in \emph{belief decomposable}
systems and extend the formulation over non-decomposable systems.
We conclude with an approximation algorithm for computing marginals,
some experimental results, and discussion.

\section{Background and Related Work}

In the flat model, a  state $s$ is a value assignment to a set of state variables, actions
$a$ affect the state through transition probabilities $\P{s'}{s,a}$, and observation tokens $o$  provide
partial information about the  resulting state $s'$ through sensing probabilities $\P{o}{s',a}$.
Given the prior $\simpleP{s}$, the target belief $\P{s}{h}$, where $h=\tup{a_0,o_0,\ldots,a_i,o_i}$
is an interleaved sequence of actions and observations, is characterized inductively as
$\P{s}{h}=\simpleP{s}$ for empty $h$, $\P{s}{h,a}= \sum_{s'}\P{s}{s',a}\!\times\!\P{s'}{h}$,
and $\P{s}{h,a,o}= \alpha\,\P{o}{s,a}\!\times\!\P{s}{h,a}$ where $\alpha=1/\P{o}{h,a}$ is a
normalizing constant.
% The posterior over any set of state variable $X$ is $P(X=x|h)=\sum_s P(s|h)$ where $s$ ranges over the states where $X=x$. 

Keeping track of beliefs using this  representation is exponential in the number of state 
variables. Approaches have thus been developed to exploit problem structure.
This structure is often made explicit through  the language  of \emph{Bayesian networks} \cite{pearl:book}.
However, while  the posterior $\P{s}{h}$ can  be obtained from a Dynamic Bayesian network (DBN) 
with as many slices as time steps \cite{dbn},   \emph{exact inference} over such networks 
is hard (yet see \cite{belgas:dbns}), so approximate inference schemes have been pursued 
instead \cite{murphy:phd}.  In the method of \citeay{bk}, 
global joint beliefs are approximated as products  of smaller local beliefs, while in particle filtering methods (PF), 
global beliefs are formed from a  set of samples \cite{kanazawa:pf,thrun:pf,koller:book}.
The  methods are related to well known approximation techniques for general Bayesian networks like 
the mini-bucket approximation \cite{mini-buckets}, (sampled) cut-set conditioning \cite{pearl:book}, 
and restricted forms of belief propagation \cite{murphy-weiss}. 

% where the functions over large set of variables that may result from variable elimination are projected 
% (and hence approximated) by smaller functions, while RB-PF methods are an approximation of 
% cut-set conditioning methods \cite{pearl:book}. 
% where values of a  select set of variables in the networks are  
% assumed to be given rendering the  network singly-connected or more generally width-bounded 
% \cite{pearl:book}. 

Our work is related to the ideas underlying these methods but it does not build explicitly on them.
It is indeed a generalization of the  beam tracking (BT) method for keeping track of  
beliefs given by \emph{sets of states} as opposed to probability distributions
\cite{bonet:jair2014}. \emph{Probabilistic beam tracking}  is aimed at combining the effectiveness of BT
% , as shown over  very large instances of problems such as Minesweeper \cite{minesweeper} and the Wumpus  \cite{russell:book}, 
with the ability to handle noisy  actions and sensing.

\section{Model, Structure, and Beams}

\noindent\textbf{Model.} 
The set of all \emph{state variables} is denoted as  $X$. A state $x$
defines a value for each state variable. In general, subsets of variables
are denoted with uppercase letters and their values with lowercase letters.
Actions $a$ affect the state stochastically with given transition
probabilities $tr(x'|x,a)$. The set of all \emph{observation variables}
is denoted with $O$ and lowercase $o$ denotes an observation; i.e.\ a value
for each of the observation variables. The sensor model is also Markovian
with probabilities $q(o|x,a)$. The joint prior is  $\simpleP{x}$.

\medskip\noindent\textbf{Histories and Beliefs.}
A history or execution $h$ is an interleaved sequence of action and
observations that begins with an action. A history is complete if it is
empty or ends with an observation. The state and observation variables
at time $t$ are denoted as $X^t$ and $O^t$ respectively. The observation
$o_t$ in the history $h$ encodes the value of the observation variables $O^t$.
% The value of the state variables $X^t$ does not form part of the histories but can be inferred from it probabilistically.
For each history $h$, either complete or incomplete, there is a probability
measure $\mathbb{P}_h$ over the events defined by the random variables
associated with $h$. We abuse notation by writing $\P{A}{h}$ instead of $\Qh{A}$, 
and $\P{A}{a,h}$ instead of $\Qsub{ha}{A}$. For a complete execution $h$
for $t$ time steps, the joint over the state variables $X^t$ is denoted
by $B^h(x)\doteq \P{x}{h}$. 
% $B^h(x)\doteq B^h(X^t=x)\doteq\P{X^t=x}{h}=\Qh{X^t=x}$. 
The \emph{prior} belief for the \emph{empty} history $h$ is $B^h(x)=\simpleP{x}$. 

\medskip\noindent\textbf{Causal Structure and 2-DBN.}
The transition and sensing probabilities are given by a 2-slice DBN whose
nodes $V$, $V'$, and $O$ stand for the state variables before and after
the action, and for the observation variables.
The transition and sensing probabilities (parameters) are
$P(V'|pa(V'),a)$ and $P(O|pa(O),a)$ where the parents $pa(V')$ of
$V'$ are among the $V$-variables, and the parents $pa(O)$ of $O$ are
among the $V'$-variables.  A variable $W$ is a \emph{cause} of $V$ if $W$ or $W'$ is a
parent of $V$ in the 2-DBN, and $W$ is \emph{causally relevant} to $V$ if it is a cause of $V$ or is causally relevant to a cause of $V$. 

\medskip\noindent\textbf{Beams.}
A \emph{beam} is a non-empty subset of state variables that is \emph{causally closed}; i.e., 
if a variable belongs to the beam, its causes must belong to the beam as well.
A \emph{collection of beams} is \emph{complete} if each state variable
is included in some beam and for each  observation variable, its set of causes
is included in some beam. The size of a beam is the number of variables that it contains,
and a complete collection of beams is \emph{minimal} if  no beam   can be replaced by a 
collection of beams of smaller size.  There is actually a unique minimal complete collection of beams 
$\cal B$ that  can be constructed as follows: if $B_i$ is the set of state variables that are causally relevant 
to a state variable $V_i$ or to an observation variable $O_i$, then $\cal B$ is the collection of beams $B_i$ 
with the  duplicate beams removed, and the  beams $B_k$ properly contained in other beams removed as well. 
By default, we assume this beam structure $\cal B$. If we enumerate the beams in  this structure as 
$B_1, \ldots, B_m$, we will refer to beam $B_i$ also by its index $i$.  This beam structure is the same 
as the one in the non-probabilistic formulation \cite{bonet:jair2014} where the \emph{causal width} of the problem 
is defined as the size of the largest beam. This structural measure is important because beam tracking runs in time 
that is \emph{exponential in the causal width}.\footnote{Actually, variables that are \emph{determined}, meaning that their initial value is known 
  and can be affected  by deterministic actions only, do not add to the causal width \cite{bonet:jair2014}.}
The same will be true for  probabilistic beam tracking. Many problems of interest can be formulated so that their causal width is bounded and small.

%% Blai: add fig as indicated below and revise par below 
\medskip\noindent\emph{Example:}
In Minesweeper, whether  in the  normal or noisy version,
there are hidden state variables $V_i$ encoding whether there is a mine at cell $i$,
and observation variables $O_i$ that after probing cell $i$ reveal the total number
of mines at the 8 surrounding cells and $V_i$ itself.
The causes of $O_i$ are the 9 cell variables $V_k$ that have no further causes
(observation variables for border cells have fewer causes).  
The minimal complete beam structure is given by beams $B_i$, one for each cell $i$,
each of size no greater than $9$. Thus, while tracking beliefs in Minesweeper is
NP-hard \cite{minesweeper}, the causal width of the problem is $9$.

\begin{figure}[t]
  \centering
  \resizebox{.975\columnwidth}{!}{
    \begin{tikzpicture}[font={\small},scale=1.0,-stealth,thick,line width=1.2,
                        a/.style={draw,minimum width=28,minimum height=25,ellipse,fill=white},
                        b/.style={draw,minimum width=28,minimum height=25,ellipse,fill=white}]
      %\node           at (0.0, 2.1) { \large Beams: };
      %\node      (b1) at (1.6, 1.8) { }; \filldraw[black] (1.6, 1.8) circle (.05); \node at (1.6, 2.1) { \large $B_1$ };
      %\node      (b2) at (3.2, 1.8) { }; \filldraw[black] (3.2, 1.8) circle (.05); \node at (3.2, 2.1) { \large $B_2$ };
      %\node      (b3) at (4.8, 1.8) { }; \filldraw[black] (4.8, 1.8) circle (.05); \node at (4.8, 2.1) { \large $B_3$ };
      %\node      (bN) at (7.4, 1.8) { }; \filldraw[black] (7.4, 1.8) circle (.05); \node at (7.4, 2.1) { \large $B_n$ };

      \node           at (8.7, 2.0) { \large Time };
      \node           at (8.7, 0.3) { \large $t$ };
      \node           at (8.7,-1.4) { \large $t+1$ };

      \node[a] (topL) at (-1.0, 0.2) { $L$ };
      \node[a] (top1) at (1.6, 0.2) { $M_1$ };
      \node[a] (top2) at (3.2, 0.2) { $M_2$ };
      \node[a] (top3) at (4.8, 0.2) { $M_3$ };
      \coordinate (top4) at (6.4, 0.2) { };
      \coordinate (top4b) at (5.8, 0.2) { };
      \node           at (6.1, 0.2) { \Large $\ldots$ };
      \node[a] (topN) at (7.4, 0.2) { $M_n$ };
      \node[a] (botL) at (-1.0,-1.5) { $L'$ };
      \node[a] (bot1) at (1.6,-1.5) { $M'_1$ };
      \node[a] (bot2) at (3.2,-1.5) { $M'_2$ };
      \node[a] (bot3) at (4.8,-1.5) { $M'_3$ };
      \node           at (6.1,-1.5) { \Large $\ldots$ };
      \node[a] (botN) at (7.4,-1.5) { $M'_n$ };
      \node[b] (obs1) at (1.6,-3.5) { $O_1$ };
      \node[b] (obs2) at (3.2,-3.5) { $O_2$ };
      \node[b] (obs3) at (4.8,-3.5) { $O_3$ };
      \coordinate (obs4) at (6.4,-3.5) { };
      \coordinate (obs4b) at (5.8,-3.5) { };
      %\node           at (4.8,-3.5) { \Large $\ldots$ };
      \node           at (6.1,-3.5) { \Large $\ldots$ };
      \node[b] (obsN) at (7.4,-3.5) { $O_n$ };

      % dynamic
      \draw (topL) -- (botL);
      \draw (top1) -- (bot1);
      \draw (top2) -- (bot2);
      \draw (top3) -- (bot3);
      \draw (topN) -- (botN);

      % obs1
      \draw (botL) -- (obs1);
      \draw (bot1) -- (obs1);

      % obs2
      \draw (botL) edge[bend right=0] (obs2);
      \draw (bot2) -- (obs2);

      % obs3
      \draw (botL) edge[bend right=0] (obs3);
      \draw (bot3) -- (obs3);

      % obsN
      \draw (botL) edge[bend right=0] (obsN);
      \draw (botN) -- (obsN);

      % beams
      %\draw[dashed,-] (b1) -- (topL);
      %\draw[dashed,-] (b1) -- (top1);
      %\draw[dashed,-] (b2) edge[bend right=10] (topL);
      %\draw[dashed,-] (b2) -- (top2);
      %\draw[dashed,-] (b3) edge[bend right=8] (topL);
      %\draw[dashed,-] (b3) -- (top3);
      %\draw[dashed,-] (bN) edge[bend right=8] (topL);
      %\draw[dashed,-] (bN) -- (topN);
      \draw[dashed,-] (topL) edge[bend left=20] (top1); \node at (0.3, 0.25) { \large $B_1$ };
      \draw[dashed,-] (topL) edge[bend left=30,out=35] (top2); \node at (2.60, 0.90) { \large $B_2$ };
      \draw[dashed,-] (topL) edge[bend left=30,out=50] (top3); \node at (4.20, 0.90) { \large $B_3$ };
      \begin{pgfinterruptboundingbox} 
        \draw[dashed,-] (topL) edge[bend left=20,out=60] (topN); \node at (6.60, 0.90) { \large $B_n$ };
      \end{pgfinterruptboundingbox} 
      %\draw (current bounding box.south east) rectangle (current bounding box.north west);
    \end{tikzpicture}
  }
  \caption{2-slice DBN and beam structure for the 1-line SLAM problem formulated with multiple
    ``dummy'' observation variables $O_i$ that are set to the value of $O$. In this model, there
    is one beam $B_i$ for each cell $i$  that  contains two variables only,  $L$ and $M_i$, for
    a causal width of $2$. 
%   By exploiting context-specific independence, this formulation becomes equivalent to the direct one. 
%   However, the new formulation has causal width 2, for any number $n$ of cells, while the causal width of the
%   direct formulation is $n+1$.
  }
  \label{fig:murphy-slam:beam:split}
\end{figure}
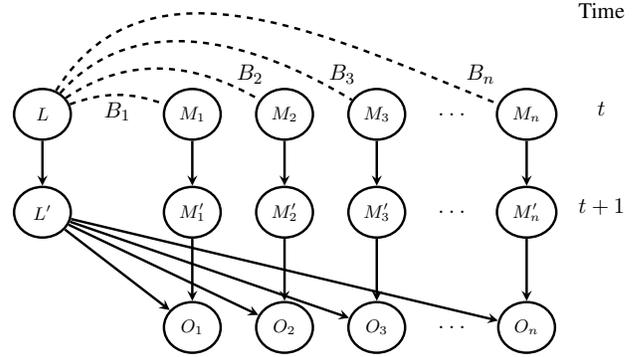

\medskip\noindent\emph{Example:}
In the direct formulation of Murphy's 1-line SLAM problem mentioned above, there are 
color variables $M_i$ for each cell $i$, an agent location variable $L$,  and one observation
variable $O$. The observation $o$ that results after applying an action $a$ is determined
by the probabilities $q(o|\ell,m_1,\ldots,m_n,a)=q(o|\ell,m_{\ell})$ that encode a form of
context-specific independence \cite{csi} in which the observation $o$  depends only  on
the color $m_{\ell}$ of the cell $\ell$  for $L=\ell$.
In this  representation, all the state variables $L$ and $\{M_i\}_i$ 
are causally relevant to the observation variable $O$, determining a beam structure with a single beam $B$ of 
size $n+1$ (Fig.~\ref{fig:murphy-slam:beam:single}). It is possible, however, to take advantage of context-specific independence 
to reformulate the model so that its \emph{causal width becomes bounded and small}.
For this, it suffices to split the single observation variable $O$ into $n$ observation
variables $O_1,\ldots,O_n$, one for each cell $i$, such that the parents of variable $O_i$
are the variables $L$ and $M_i$ only, and to set the probability $q_i(o_i|\ell,m_i)$ for
variable $O_i$ equal to $q(o_{\ell}|\ell,m_{\ell})$ when $\ell=i$ and to $1/2$ (a normalizing constant) when $\ell\neq i$. 
The value of all the ``artificial'' observation variables  $O_i$ at time $t$ is set to the
value of the real observation variable $O$ at time $t$; that is, if the observation
$O$ has value $o$ at time $t$ then all variables $O_i$ are set to $o$ at time $t$.
The sensor model of the reformulated task is defined as
$q(\tup{o_1,\ldots,o_n}|\ell,m_1,\ldots,m_n,a)=\prod_{i=1}^n q_i(o_i|\ell,m_i)$.
While the two models are equivalent, the first has one beam of size $n+1$, while the
latter has $n$ beams $B_i$ of size $2$, each containing the agent location variable $L$
and the cell variable $M_i$, $i=1, \ldots, n$, for a \emph{causal width} of $2$ (Fig.~\ref{fig:murphy-slam:beam:split}).
%The two different 2-DBN  models for the problem are shown in Figure~\ref{fig:1-slam} along with the resulting beams.  
%When referring to the model for  1-line SLAM and similar  problems, it's the latter type of model that we have in mind. 

\begin{figure}[t]
  \centering
  \resizebox{.9\columnwidth}{!}{
    \begin{tikzpicture}[font={\small},scale=1.0,-stealth,thick,line width=1.2,
                        a/.style={draw,minimum width=28,minimum height=25,ellipse}]
      \node[a] (topL) at (0.0, 0.2) { $L$ };
      \node[a] (top1) at (1.6, 0.2) { $M_1$ };
      \node[a] (top2) at (3.2, 0.2) { $M_2$ };
      \node           at (4.5, 0.2) { \Large $\ldots$ };
      \node[a] (topN) at (5.8, 0.2) { $M_n$ };
      \node[a] (botL) at (0.0,-1.5) { $L'$ };
      \node[a] (bot1) at (1.6,-1.5) { $M'_1$ };
      \node[a] (bot2) at (3.2,-1.5) { $M'_2$ };
      \node           at (4.5,-1.5) { \Large $\ldots$ };
      \node[a] (botN) at (5.8,-1.5) { $M'_n$ };
      \node[a]  (obs) at (3.2,-3.0) { $O$ };
      \draw (topL) -- (botL);
      \draw (top1) -- (bot1);
      \draw (top2) -- (bot2);
      \draw (topN) -- (botN);
      \draw (botL) edge[bend right=10] (obs);
      \draw (bot1) -- (obs);
      \draw (bot2) -- (obs);
      \draw (botN) -- (obs);
      \draw[dashed] (2.9, 0.2) ellipse (4.2cm and .8cm);
      %\draw[dashed] (3.0,-1.5) ellipse (4.5cm and .7cm);
      \node           at (3.2, 1.4) { \large Beam $B$ };
      \node           at (8.0, 1.4) {\large Time };
      \node           at (8.0, 0.3) {\large $t$ };
      \node           at (8.0,-1.4) {\large $t+1$ };
    \end{tikzpicture}
  }
  \caption{2-slice DBN and beam structure for the direct formulation of the 1-line SLAM problem. % with a single observation variable $O$.
    The observation probabilities satisfy $\P{o}{\ell,m_1,\ldots,m_n,a}=q(o|\ell,m_{\ell})$ that
    corresponds to a form of context-specific independence.
    This independence is not exploited and results in a beam structure that contains
    a single beam $B$ with all the $n+1$ state variables.
  }
  \label{fig:murphy-slam:beam:single}
\end{figure}
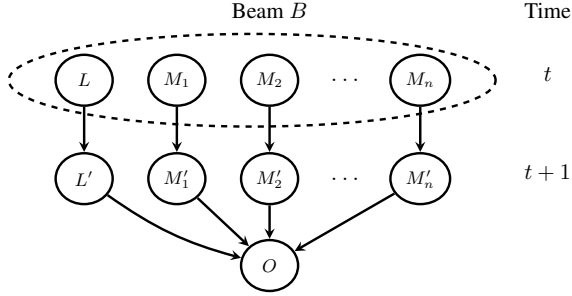

\medskip\noindent\textbf{Internal and External Variables.} A variable that appears in more than one beam is called  \emph{external}, 
while one that appears in one beam only  is called  \emph{internal}. The internal and external variables for beam $X_j$ are denoted as $Y_j$ and $Z_j$ respectively.
The set $X$ of all (state) variables is partitioned as $X=YZ$ where $Y$ are all the internal  variables and $Z$ are all the external variables.
In the second formulation of 1-line SLAM, the color variables are all internal, and the agent location variable is external. 
In Minesweeper, all variables are external. 

\medskip\noindent\textbf{Factored Model}. 
Given the beam structure determined by the 2-DBN structure, the transition
and sensing probabilities $tr(x'|x,a)$ and $q(o'|x',a)$ can be factorized
as $tr(x'|x,a) = tr(y'z'|x,a)=tr(y'|z',x,a) tr(z'|x,a)$ with
$tr(y'|z',x,a)=\prod_j tr_j(y'_j|z'_j,x_j,a)$, 
$tr(z'|x,a)=\prod_j tr_j(z'_j|x_j,a)$, and
$q(o'|x',a)=\prod_j q_j(o'_j|x'_j,a)$, where $j$ ranges over the  beam indices. % , and the sensing probabilities as $q(o'|x',a)=\prod_j q_j(o'_j|x'_j,a)$. 
All the $tr_j$ and $q_j$ probabilities are determined by the conditional
probabilities $P(V'|pa(V'),a)$ in the 2-DBN.\footnote{If $Y_j$ is $\{V_1, \ldots, V_k\}$
  where the variables $V_i$ are ordered topologically, $tr_j(y'_j|z'_j,x_j,a)$
  is $\smash[t]{\prod_{i=1}^k} P(V'_i|pa(V'_i),a)$. Associating each external variable $V$
  with the smallest $j$ such that $B_j$ contains $V$, and each observation 
  variable $W$ with the smallest $j$ such that $B_j$ contains all of its causes,
  $tr_j(z'_j|x_j,a)$ factorizes as $\smash[t]{\prod_{i=1}^k} P(V'_i|pa(V'_i),a)$, where $V_1, \ldots, V_k$ are the external
  variables in $Z_j$ associated with the beam $j$ in topological order, and
  $q_j(w'|x'_j,a)$ is $P(w'|pa(w'),a)$ when the observation variable $W$ is
  associated with the beam $j$, and else $q_j(w'|x'_j,a)=1$.}
It is also assumed without loss of generality that the prior belief $B^h(x)$
for the empty history factorizes across the beams as $\prod_j B^h_j(x_j)$.
When this is not the case, the model can be extended with an extra action-observation
pair that must start any non-empty history.

\section{Tracking in Belief Decomposable Systems}

Given the model structure and beams, our task is to show that the posterior
joint beliefs $B^h(x)= B^h(X^t=x)$ for  histories $h=\tup{a_0,o_0,\ldots,a_{t},o_{t}}$
can be expressed as the normalized product of  \emph{belief factors} $\prod_j B^h_j(x_j)$, one for 
each beam $j$, that are tracked independently. For this, we assume a form of \emph{determinism} over 
the state variables that appear in more than one beam; namely, the external variables.
%Later on, we will see that when this critical assumption does not hold, it can be enforced by sampling.
The following key definition is from \citeay{bonet:jair2014}:

\begin{definition}[Backward Determinism]
\label{def:BD}
A state variable $V$ is \emph{backward deterministic} if the value of $V$ at any time $t$
is determined by the value of\, $V$ at time $t+1$, the action at time $t$, the history $h$
up to time $t$, and the priors. 
\end{definition}

Notice that static variables and variables that are initially fully known and
are affected by deterministic actions only  are backward deterministic, as  are the
variables that are fully observable  and the variables $V$ affected by deterministic 
actions that map different values of $V$ into different values of $V'$ \cite{amir:filtering}.
When $V$ is  backward deterministic, we write $\R_a(v|h)$ to denote the
value of $V$ at time $t$ that is determined by the value $v$ of $V$ at time $t+1$,
the complete history $h$ up to time $t$, the next action $a$, and the priors.
We refer to $\R_a(v|h)$ as the \emph{regression} of $V=v$ given $h$, $a$, and the
priors (left implicit in the notation).
%A model is \emph{belief decomposable} when all the external variables are backward deterministic:

\begin{definition}[Belief Decomposable Model]
\label{def:decomposable-model}
A model is \emph{belief decomposable} when all the external variables are backward deterministic.
\end{definition}

It will be convenient to  use the abbreviation BD to refer to both \emph{belief decomposable models}
and to \emph{backward deterministic} variables. The notation $\R_a(z|h)$ for the set $Z$ of external
BD variables  denotes the regression of the value vector $z$ through $h$ and $a$.
If the value $v$ for a variable or set of BD variables $V$ is impossible given $h$ and $a$, we write
$\R_a(z|h)=\bot$.
%  We assume that the regression $\R_a(z|h)$ factorizes across the beams as $\R_a(z|h)_j=\R_a(z_j|h)$;
% that is, we assume that each beam contains enough information to compute regressions of the
% external variables in the beam.
Clearly, $\R_a(z|h)=\bot$ iff $\R_a(z_j|h)=\bot$ for some beam $j$, and $\R_a(z|h)_j=\R_a(z_j|h)$.
% Consequently, the indicator function $I(z)=\I{\R_a(z|h)\neq\bot}$ decomposes as  $\prod_j \I{\R^j_a(z_j|h)\neq\bot}$.
% \Alert{(THIS IS AN ASSUMPTION ABOUT THE BEAMS)} 
% no assumption here .. follows from beams causally closed and def of BD variables .. 

\subsection{Equations for the Belief Factors}

We want to show that the  distribution $B^h(x) = \P{x}{h}$ after history
$h$ for a BD model is the normalized product of factors  $B^h_j(x_j)$:
\begin{equation}
B^h(x) \ = \ \textstyle\beta \, \prod_j B^h_j(x_j)
\label{eq:factors}
\end{equation}
where $B^h_j(x_j)$ denotes the belief factor over the  variables in
beam $j$, and $\beta=\beta(h)$ is a normalization factor that only depends on $h$.
We show this inductively by using the assumption of factorized priors and by
expressing the factors that define the joint belief $B^{h'}(x)$ for $h'= \tup{h,a,o}$
in terms of the factors that define the joint belief $B^{h}(x)$, where $a$
is an action and $o$ an observation such that $P(o|a,h) > 0$. 
The factors $B^h_j(x_j)$ for the empty history $h$ are given.

\medskip\noindent
Let $x'=y'z'$ be a valuation for $X^{t+1}$. Using Bayes' rule, the posterior can be expressed as
\begin{alignat}{1}
  \P{x'}{o,a,h}   
% &\ =\ \alpha\, \P{x',o}{a,h} \\
  &\ =\ \alpha\, \P{o}{x',a,h}\,\P{x'}{a,h} 
\label{eq:complete:0}
\end{alignat}
where $\alpha=1/\P{o}{a,h}$ is a normalizing constant,
and the second term is
\begin{alignat}{1}
  \P{x'}{a,h} 
% &\ =\ \textstyle\sum_y \P{x',y}{a,h} \\
%   &\ =\  \textstyle\sum_y \P{y',z',y}{a,h} \\
\label{eq:complete:1}
  &\ =\ \textstyle\sum_y \P{y'}{z',y,a,h}\,\P{z',y}{a,h} \,.
\end{alignat}
Assume now that $\R_a(z'|h)\neq\bot$. Using backward determinism and 
factored transitions, the first term in \eqref{eq:complete:1} becomes:
\begin{alignat}{1}
  \P{y'}{z',y,a,h}\
% \label{eq:complete:2}
%       &=\ \P{y'}{z',y,\R_a(z'|h),a,h} \\
 \label{eq:complete:4}
       &=\ tr(y'|z',y,\R_a(z'|h),a) \\
\label{eq:complete:6}
      &=\ \textstyle\prod_j tr_j(y'_j|z'_j,y_j,\R_a(z'|h)_j,a) \,.
\end{alignat}
For the second term, we use the inductive hypothesis:
%For the second term in \eqref{eq:complete:1}, we use the inductive
%hypothesis in \eqref{eq:complete:10} below to obtain:
\begin{alignat}{1}
  &\P{z',y}{a,h} 
     \label{eq:complete:7}
   \,=\ \P{z',y,\R_a(z'|h)}{a,h} \\
\label{eq:complete:8}
  &\,=\ \P{z'}{y,\R_a(z'|h),a,h}\ \P{y,\R_a(z'|h)}{a,h} \\
\label{eq:complete:9}
  &\,=\ tr(z'|y,\R_a(z'|h),a)\ \P{y,\R_a(z'|h)}{h} \\
\label{eq:complete:10}
  &\,=\ tr(z'|y,\R_a(z'|h),a)\  \beta\, \textstyle\prod_j B^h_j(y_j,\R_a(z'|h)_j) \\
\label{eq:complete:11}
  &\,=\ \beta\textstyle\prod_j tr_j(z'_j|y_j,\R_a(z'_j|h),a)\,B^h_j(y_j,\R_a(z'_j|h))\,.
\end{alignat}
Substituting these expressions back into \eqref{eq:complete:1},
abbreviating $\smash[b]{\R_a(z'_j|h)}$ as $\smash[b]{\R(z'_j)}$, and using $Y_i\cap Y_j=\emptyset$ for $i\neq j$:
% ,  using $Y_i\cap Y_j=\emptyset$ for
% $i\neq j$ in \eqref{eq:complete:disjoint},  factored backward determinism in
% \eqref{eq:complete:fbd1}, 
%
\begin{alignat}{1}
  &\P{x'}{a,h} \, \beta^{-1} \ = \nonumber \\ 
%       &\quad=\ \alpha \underbrace{ \textstyle\prod_j q_j(o_j|x'_j,a) }_{\P{o}{x',a,h}} \,
%                       \textstyle\sum_y
%                           \underbrace{ \textstyle\prod_j tr_j(y_j'|z'_j,y_j,\R_a(z'|h)_j,a) }_{\P{y'}{z',y,a,h}} \,
%                           \underbrace{ \beta \textstyle\prod_j
%                                             tr_j(z'_j|y_j,\R_a(z'|h)_j,a) \,
%                                             B^h_j(y_j,\R_a(z'|h)_j) }_{\P{z',y}{a,h}} \\
%      &\quad=\ \beta\, \textstyle\sum_y \prod_j
%                                 q_j(o_j|x'_j,a)\,
%                                 tr_j(y_j'|z'_j,y_j,\R_a(z'|h)_j,a)\,
%                                 tr_j(z'_j|y_j,\R_a(z'|h)_j,a)\,
%                                 B^h_j(y_j,\R_a(z'|h)_j) \\
      &=\ \textstyle\sum_y \prod_j
%                                 q_j(o_j|x'_j,a)\,
                                 tr_j(y_j',z'_j|y_j,\R(z'_j),a)\, % \R_a(z'|h)_j,a)\,
                                 B^h_j(y_j,\R(z'_j)) \\
    \label{eq:complete:disjoint}
      &=\ \textstyle\prod_j
 %                                q_j(o_j|x'_j,a)\,
                                 \sum_{y_j}
                                     tr_j(y'_j,z'_j|y_j,\R(z'_j),a)\, %\R_a(z'|h)_j,a)\,
                                     B^h_j(y_j,\R(z'_j))  % \R_a(z'|h)_j) 
    \\
    \label{eq:complete:fbd1}
      &=\ \textstyle\prod_j
%                                  q_j(o_j|x'_j,a)\,
                                 \sum_{y_j}
                                     tr_j(x'_j|y_j,\R(z'_j),a)\, % \R^j_a(z'_j|h),a)\,
                                     B^h_j(y_j,\R(z'_j)) \,. % \R^j_a(z'_j|h))
\end{alignat}
Finally, from the factorization of the observations
\begin{alignat}{1}
  \P{o}{x',a,h}\ &=\ q(o|x',a)\ =\ \textstyle\prod_j q_j(o_j|x'_j,a) \,,
\end{alignat}
and the inductive hypothesis, %induction $\P{x'}{o,a,h} =  B^{h'}(x')$,
% \alpha\beta\, 
 %  \I{\R_a(z'|h)\neq\bot}
%                       \textstyle\prod_j
%                           q_j(o_j|x'_j,a)\,
%                           \sum_{y_j}
%                               tr_j(y'_j,z'_j|y_j,\R^j_a(z'_j|h),a)\,
%                               B^h_j(y_j,\R^j_a(z'_j|h)) 
%   \label{eq:complete:fbd2}
% \\  &=\ \beta'\, \textstyle\prod_j
%                          \I{\R^j_a(z'_j|h)\neq\bot}\,
%                           q_j(o_j|x'_j,a)\, 
%  \sum_{y_j}     tr_j(y'_j,z'_j|y_j,\R(z'_j),a)\, % \R^j_a(z'_j|h),a)\,
%                               B^h_j(y_j,\R(z'_j)) % \R^j_a(z'_j|h))
%     =\ \beta\, \textstyle\prod_j B^{h'}_j(x'_j)
% \end{alignat}
the factors $B^{\smash[t]{h'}}(x'_j)$ become:
\begin{alignat}{1}
  & \smash[t]{B^{h'}_j(y'_j,z'_j)}\ 
     =\ % \I{\R^j_a(z'_j|h)\neq\bot}
 \alpha'   \, \smash{q_j(o_j|y'_j,z'_j,a)} \ \times \ \nonumber \\
 & \quad \textstyle\smash{\sum_{y_j} tr_j(x'_j \, | \, y_j,\R_a(z'_j|h),a)\, B^h_j(y_j,\R_a(z'_j|h))} 
   \label{eq:complete:cuasi}
\end{alignat}
when $\R_a(z'|h)\neq\bot$, and $B^{h'}_j(y'_j,z'_j)=0$ otherwise.
That is, the  belief factors $B_j^h$ are progressed and filtered \emph{independently} for each beam $j$. 
The complexity of updating each belief factor, i.e., mapping $B^h_j$ into $B^{h'}_j$ for $h'=\tup{h,a,o}$,
is exponential in the  beam $j$ size, and more precisely, in the number of internal  variables  in  beam $j$.

\Omit
{These factors in Equation~\ref{eq:complete:fbd2} along with the factors representing the prior distribution, 
legitimate the decomposition captured by Equation~\ref{eq:factor} where the global belief $B^h$ is 
captured as the normalized product of the local factors $B^h_j$. The factorization also defines
a simple belief tracking algorithm that will be efficient if the beams are small, and  will be exact
if the backward determinism assumptions hold for the variables appearing in more than one beam. 
}

While the factors $B^h_j$ determine the joint distribution $B^h$,
the computation of marginals from such a joint is hard in general.
We will come back to this issue but  focus now  on extending the formulation
to models where not all external variables are backward deterministic.
For this, like in Rao-Blackwellized PF methods, we sample such variables
to make them backward deterministic given their sampled history.\footnote{%
  The assumption of backward determinism appears in \eqref{eq:complete:cuasi} through the use of 
  the regressions $\R_a(z'_j|h)$. A different type of approximation can be defined by replacing
  such regressions by values $z_j$ that are summed over with weights given by $\P{z_j}{z'_j,h,a}$.
  When the BD assumption holds, these weights are either $0$ or $1$, and the new formula reduces
  to the old formula. When BD does not hold, the new formula encodes an approximation. 
  We tested this approximation empirically, but in the examples considered, it does not
  run faster than the particle-based approximation  and its quality does not appear 
  to be better either.}

\section{Tracking in Non-Decomposable Systems}

We assume now that the set $X$ of state variables is partitioned
into three sets $Y\!ZU$ where $Y$ stands for the internal variables
(variables appearing in one beam only), $Z$ for the external
variables that are backward deterministic, and the new set $U$ for the external variables
that are not BD and need to be sampled. 
If the context above was provided by the history $h$ of actions and
observations, the new context is provided by $h$ and the sampled
history $\bar u$ of the $U$ variables.
The expression $\R_a(z',u'|\bar u',h)$ denotes the pair of values
$\R_a(z'|h)$ and $\R_a(u'|\bar u')$ where the latter denotes $u$,
the value preceding the last value $u'$ in the sampled history $\bar u'$. 
The joint over the $Y$ and $Z$ variables can be expressed as 
\begin{equation}
\P{y,z}{h}\ =\ \textstyle\sum_{\bar u} \P{y,z}{\bar u,h}\,\P{\bar u}{h}
\label{eq:sampling1}
\end{equation}
which can be approximated as:
\begin{equation}
\P{y,z}{h}\ \approx\ \textstyle\sum_{i=1}^N \P{y,z}{\bar u_i,h}
\end{equation}
where the histories $\bar u_i$ are sampled with probability $\P{\bar u_i}{h}$.
It is often convenient however to sample $\bar u_i$ with a different
distribution $\pi({\bar u}|h)$, called the \emph{sampling} or
\emph{proposal distribution}, as long as no possible $U$ history is
made impossible. %; i.e., provided that  $\pi({\bar u}|h) = 0$ implies $\P{\bar u}{h} = 0$.
In this method, called \emph{importance-based sampling} \cite{bishop:book} 
the approximation becomes: 
\begin{equation}
\P{y,z}{h}\ \approx\  \alpha\,\textstyle\sum_{i=1}^N w_i \times \P{y,z}{\bar u_i,h} 
\label{eq:sampling-approx}
\end{equation}
where $\alpha$ is a normalizing constant, and the weights $w_i$ are
\begin{equation}
w_i\ =\ \P{\bar u_i}{h}/\pi({\bar u_i}|h) \,.
\label{eq:weights}
\end{equation}
Provided with the sampled histories $\bar u$, it can be shown that
$\P{y,z}{\bar u,h}$ in \eqref{eq:sampling1} becomes:
\begin{equation}
\label{eq:filter:factor}
\P{y,z}{\bar u,h}\ 
  =\  B^h(y,z|\bar u)\ 
  =\ \beta\,\textstyle\prod_j B^h_{j}(y_j,z_j|\bar u) 
\end{equation}
where the factors $B^h_{j}(\,\cdot\,|\,\bar u)$ can be progressed
to $h'=\tup{h,a,o}$ and $\bar u'=\tup{\bar u,u'}$ as:
\begin{alignat}{1}
& B^{h'}_j(y'_j,z'_j|\bar u')\ =\  \alpha'' \, q_j(o_j|x'_j,a)\ \times \nonumber \\
& \ \textstyle\sum_{y_j}  tr_j(x'_j|y_j,\R_a(z'_j|h),u_j,a)\,B^h_j(y_j,\R_a(z'_j|h)|\bar u) 
\label{eq:filterj}
\end{alignat}
when $\R_a(z'_j|h) \neq\bot$, else $B^{h'}_j(y'_j,z'_j|\bar u')=0$.
This equation is indeed exactly like \eqref{eq:complete:cuasi} except for
the sampled history $\bar u$ included in the context, and the new
component $u'_j$  in $x'_j$.

In summary, the filter for approximating the target joint distribution $\P{y,z}{h}$
is given by the sequence of triplets  $\F_h=\{(\bar u_i, w_i, \{B^h_{j}\}_j)\}_i$
where $h$ is the action-observation history, $\bar u_i$ is the sampled history of
the $U$ variables, $i=1, \ldots, N$, $w_i$ is the weight associated with $\bar u_i$,
and $B^h_{j}(y_j,z_j|\bar u_i)$ represents the belief factor given $h$ and $\bar u_i$
for each beam $j$.
These  belief factors determine the probability $\P{y,z}{\bar u,h}$ via \eqref{eq:filter:factor},
that provides the approximation for $\P{y,z}{h}$  via \eqref{eq:sampling-approx}.
The filter $\F_h$ is extended to $\F_{h'}$ for $h'=\tup{h,a,o}$ by 1)~extending
the sampled history, 2)~computing its associated weight, and 3)~updating the belief factors
with the new action-observation pair and the new sample.
This last operation is defined by \eqref{eq:filterj}.
We focus next on the other two operations.
Initially,  $\bar u_i$ is empty and $w_i=1$ for $i=1, \ldots, N$. 

\subsection{Marginals, Samples, and Weights}

The factors $B^h_{j}(x_j|\bar u)$, where $x_j$ represents 
the valuations $y_jz_j$ over the non-sampled variables in beam $b_j$, 
do not represent themselves the probabilities $\P{x_j}{h,\bar u}$, 
that stand actually for the marginals:
\begin{alignat}{1}
\P{x_j}{h,\bar u}\ &=\ \beta\,\textstyle\sum_{w} \prod_i B^h_{i}(x_i|\bar u)
\label{marginals}
\end{alignat}
where $\beta$ is a normalizing constant and the sum ranges over all the variables $W$ that are not in the beam $j$. 
Such marginals can be computed from the factors $B^h_{j}(x_j|\bar u)$ by
standard methods  such as the jointree algorithm or belief propagation.
%For such a computation, each beam $j$ is identified with a meta-variable
%$X_j$ whose possible values correspond to the valuations of the variables
%in the beam. 
We  show next  how to compute the proposal distribution and weights from
such   marginals. 

We consider two proposal distributions $\pi$: the so-called \emph{motion}
and \emph{optimal} distributions \cite{doucet:slam,burgard:slam}.
The motion distribution uses only the transition probabilities to
generate new histories $\bar u'=\tup{\bar u,u'}$ from previous
histories $\bar u$ (not using the information provided by $o$):
\begin{alignat}{1}
\pi_{motion}(u'|h,a,o,\bar u)\ &=\ \P{u'}{h,a,\bar u} \,,
\end{alignat}
while the optimal proposal makes use of both $a$ and $o$:
\begin{alignat}{1}
\pi_{opt}(u'|h,a,o,\bar u)\ &=\ \P{u'}{h,a,o,\bar u} \,.
\end{alignat}

For computing these probabilities effectively, we will \emph{assume}
that there is at least one beam $j$ that contains the whole set of
variables $U$, i.e. for which $U = U_j$.
This assumption holds automatically when $U$ is a singleton as in
many SLAM problems. In addition and without loss of generality,
we assume that the observation $o=o^t$ at time $t$, for each time step,
falls into a single beam only; i.e., $q(o|x,a)=q_j(o_j|x_j,a)$ for
some beam $j$, and $q_k(o_k|x_k,a)=1$ for $k \neq j$.
This is true when one observation variable is observed at each time point,
and when this is not true, it can be made true by serializing the
observations, using dummy actions in between.
Under these assumptions, the proposal distributions and their weights
can be computed from the marginals $P(x_j|h,\bar u)$ associated with
one beam. Indeed, if $U=U_j$ and $x_j=y_j,z_j,u_j$:
\begin{alignat}{1}
 \P{u'}{h,a,\bar u}\ &=\ \textstyle\sum_{y_j,z_j} \P{u'}{x_j,a} \, \P{y_j,z_j}{h,\bar u}
\label{pi-motion}
\end{alignat}
and $\P{u'}{h,a,o,\bar u} \propto \P{o}{h,a,\bar u'}\P{u'}{h,a,\bar u}$ where 
\begin{alignat}{1}
\P{o}{h,a,\bar u'}\ &=\ \textstyle\sum_{y'_k,z'_k} q_k(o_k|x'_k,a) \, \P{x'_k}{h,a,\bar u'} 
\end{alignat}
if $o$ falls into the  beam $k$. The marginal $\P{x'_k}{h,a,\bar u'} = \P{y'_k,z'_k}{h,a,\bar u'}$
excludes the observation $o$, and hence it is obtained with
\eqref{eq:filterj} by setting all $q_j(o_j|x'_j,a)$ to $1$. 

The weights $w' = \P{\bar u'}{h'}/\pi(\bar u'|h')$ can be computed incrementally from the 
weights $w = \P{\bar u}{h}/\pi(\bar u|h)$ when $h' = \tup{h,a,o}$ and $\bar u' = \tup{\bar u,u'}$.
For the motion proposal, the weight is the ratio between:
\begin{alignat}{1}
\P{\bar u'}{h'}\  &=\ \alpha'\,\P{o}{\bar u',h,a}\,\P{u'}{h,a,\bar u}\,\P{\bar u}{h}  \nonumber 
\label{sampling}
\end{alignat}
and $\pi_{motion}(\bar u'|h') = \P{u'}{h,a,\bar u} \pi(\bar u|h)$,  which results in:
\begin{alignat}{1}
w'_{motion}\ &=\ \alpha'\,w_{motion} \times \P{o}{\bar u',h,a}  \,.
\end{alignat}
For the optimal proposal, the weight is the ratio between 
\begin{alignat}{1}
\P{\bar u'}{h'}\ &=\ \alpha''\,\P{u'}{h,a,o,\bar u}\,\P{o}{h,a,\bar u}\,\P{\bar u}{h} \nonumber
\end{alignat}
and $\pi_{opt}(\bar u'|h') = \P{u'}{h,a,o,\bar u} \pi(\bar u|h)$ which results in:
\begin{alignat}{1}
w'_{opt}\ &=\ \alpha''\,w_{opt} \times \P{o}{h,a,\bar u} \,.
\end{alignat}
If the observation $o$ falls into beam $k$, the  marginals required can be computed as: 
\begin{alignat*}{1}
&\P{o}{\bar u',h,a}\ =\ \textstyle\sum_{x'_k} q_k(o|x'_k,a) \, \P{x'_k}{h,a,\bar u'}  \,\text{, \ and} \\
&\P{o}{\bar u,h,a}\ = \\
&\quad\qquad\textstyle\sum_{x'_k} q_k(o|x'_k,a)\sum_{u'}\P{x'_k}{h,a,\bar u'}\P{u'}{h,a,\bar u}
\end{alignat*}
%and $\P{o}{\bar u,h,a}=\sum_{x'_j} q_j(o|x'_j,a)\P{x'_j}{h,a,\bar u'}\P{u'}{h,a,\bar u}$, 
where $\P{x'_j}{h,a,\bar u'} = \P{y'_j,z'_j}{h,a,\bar u'}$  is obtained as indicated above,
and $\P{u'}{h,a,\bar u}$ is given by \eqref{pi-motion}.

\section{Faster Approximation of Marginals}

The main bottleneck of the  algorithm is the computation of the marginals
$\P{x_j}{h,\bar u}$ from the factors $B^h_j(x_j|\bar u)$ following \eqref{marginals}.
Such marginals are needed for computing the samples and weights, and
for answering queries. The marginals can be computed using the 
jointree algorithm or belief propagation (BP). However, since  scalability 
is crucial, we  introduce a third method that will be evaluated in comparison with the other two.
It is motivated by the results reported by beam tracking that uses arc consistency (AC)
\cite{ac,dechter:book} for progressing logical (non-probabilistic) beliefs.
The relation between BP and AC is well-known: both methods are exact for trees and BP
propagates the zero-probability entries in agreement with AC \cite{ac-bp}. 
We call the new method Iterated AC which is aimed at combining the speed and monotonic
convergence of AC with the ability to approximate probabilistic beliefs, 
even if roughly.
% While the method is new, it is based on well known ideas and will provide a useful
% reference point to assess the quality-speed tradeoff; in particular, for cases
% when the distributions defining the model are peaked, a situation
% that is common in SLAM \cite{slam:peaked} and other settings.

Iterated AC approximates the marginals from the belief factors 
by using arc consistency  along with  order-of-magnitude probabilities also called $\kappa$-rankings
\cite{spohn:kappas,pearl:kappas,darwiche1994relation}. For this, it follows  three steps 
that we sketch briefly with no much justification.
The algorithm uses two parameters $\epsilon$ and $m$. 
\emph{First}, real values $p\in(0,1]$ are mapped into the smallest non-negative integer
$\kappa=\kappa_\epsilon(p)$ for which  $\epsilon^{\kappa+1} < p$, while  $p=0$ is mapped
into $\kappa_\epsilon(p)=\infty$.
% E.g., if $\epsilon=0.2$, then values of $p$ equal to $0.4$, $0.1$, and $0.01$
% are mapped into $\kappa_\epsilon(p)$ values of  $0$, $1$, and $2$ respectively.
% If $p$ is a probability and $\epsilon$ is sufficiently small, $\kappa=\kappa_\epsilon(p)$ 
% represents roughly the ``order-of-magnitude'' of $p$.
%%
This mapping is used to transform the belief factor $B^h_j(\,\cdot\,|\bar u)$ into tables 
$D^i_j$  that contain all  the valuations $x_j$ (for the variables $X_j$
in beam $j$) that satisfy $\kappa_\epsilon(B^h_j(x_j|\bar u)) \leq i + \eta_j$
for a given non-negative integer $i$ where $\eta_j=\min_{x_j}\kappa_\epsilon(B^h_j(x_j|\bar u))$ is a  normalization constant.
% That is, $D^i_j$ contains all valuations $x_j$ for the beam $j$ that 
% ``seem to be likely enough'' according to $\epsilon$,  $i$, and 
% the factors  $B^h_j(x_j|\bar u)$.
%%
\emph{Second}, the tables $D_j^i$ associated with the different beams $j$
and the same $i$  are made \emph{arc consistent} (the tables share
variables). The $\kappa$-marginal $\kappa(x_j|h,\bar u)$ is 
defined then  as $i$ iff $i$ is the minimum non-negative integer for which the 
tuple $x_j$ belongs to  $D_j^i$ after running AC, while  $\kappa(x_j|h,\bar u)=\infty$ when there is no such $i$.
%$\kappa(x_j|h,\bar u)$ is set to $\infty$.
%marginal $\kappa(x_j|h,\bar u)$ values over the beam $j$, the histories $h$ and $\bar u$, 
%are set to $i$ iff $i$ is the min value for which the value $x_j$ for the  table $D^{h,i}_{\bar u,j}$ 
%is \emph{arc consistent} with the tables $D^{h,i}_{\bar u,k}$ for the other  beams $k \not= j$. 
%That is, we check, approximately, if there is a joint valuation $x$ that extends $x_j$
%such that $x_k$ belongs to each table $D_k$ for all beams $k$. 
%%
\emph{Finally}, the $\kappa(x_j|h,\bar u)$ marginals are used to 
approximate the  marginals as $\P{x_j}{h,\bar u}= \alpha \epsilon^{\kappa(x_j|h,\bar u)}$
where $\alpha$ is a normalizing constant. 

A further simplification is that $\kappa$ measures that
are greater than the $m$ parameter but less than $\infty$, are  treated as if 
they were equal to $m$. This means that the approximation of the 
marginals $\P{x_j}{h,\bar u}$ from the belief factors $B^h_j(\,\cdot\,|\bar u)$
are computed  by running arc consistency $m+1$ times.
Iterated AC will be denoted as \ACm\ where $m$ is the parameter used.
The parameter $\epsilon$ is fixed to $0.1$.

\section{Experimental Results}

The  general PBT algorithm can use different  algorithms for
computing marginals from the factors. We experiment
with the  jointree (JT), belief propagation (BP), and \ACm\
algorithms for $m \in \{0,1\}$.
For JT and BP we use \texttt{libdai} \cite{Mooij_libDAI_10}
while \ACm\ is ours.
The experiments were performed on Intel Xeon E5-2666 CPUs running
at 2.9GHz with a memory cap of 10Gb (exhausted/approached only by JT).

\pagebreak
\medskip\noindent
\textbf{Minesweeper.}  Minesweeper cannot be fully solved by
pure inference and requires guessing in certain situations,
even if variables are static and sensing is noiseless. 
Since all  variables are external but static, and hence backward deterministic,
no sampling is required.  The beam decomposition has one
beam $B_i$ for each cell $i$ that contains up to 9
variables. Table~\ref{table:ms} shows the results for PBT
using JT, BP, and \ACzero, with a policy that chooses
to tag (resp.\ open) the cell that is most certain to 
contain a mine (resp.\ to be clear).
The success ratio indicates the percentage of 
maps solved, i.e., without doing a wrong action.
In this noise-free example, it can be shown that the 
marginals computed by PBT with \ACzero\ are  equivalent 
to the ones computed by beam tracking where the 
belief factors represent sets of states \cite{bonet:jair2014}.
Actually,  PBT with \ACzero\ scales up best with 
a quality that matches the quality of JT in the small
instances. JT does not scale up to larger instances
and BP does but achieves a  much lower score.

\begin{table}
\centering
\resizebox{\columnwidth}{!}{
\begin{tabular}{@{}crrrrrrr@{}}
  \toprule 
        &       & \multicolumn{2}{c}{JT} & \multicolumn{2}{c}{BP} & \multicolumn{2}{c}{\ACzero} \\
  \cmidrule(lr){3-4}
  \cmidrule(lr){5-6}
  \cmidrule(l){7-8}
         board & mines & \%succ &  time & \%succ &  time & \%succ &  time \\
  \midrule
    6$\times$6 &     6 &   84.1 & .002 &   68.5 & .046 &   84.2 & .000 \\
    8$\times$8 &    10 &   83.3 & .070 &   66.3 & .069 &   84.6 & .002 \\    
  16$\times$16 &    40 &    --- &  --- &   41.4 & .232 &   79.9 & .005 \\
  30$\times$16 &    99 &    --- &  --- &    1.8 & .991 &   33.4 & .003 \\
  \bottomrule
\end{tabular}
}
\caption{PBT in Minesweeper using three methods for computing marginals. 
  Time is average time per decision in seconds.
  JT runs out of memory in larger maps. Figures are  averages over 500 runs.
}
\label{table:ms}
\end{table}

\Omit{ I'm removing due to lack of space ... if space is made ..
\medskip\noindent
\textbf{Color-tile SLAM.} This is a 2-dimensional SLAM
task where observations depend on the color of the current
cell only. In this case, PBT reduces to the well-known Rao-Blackwellized
PF in which all beams are pairwise disjoint and inference
become tractable. For space reasons, the reader is referred to
\cite{murphy:1999} for formal and experimental results.
}

\medskip\noindent
\textbf{1-line-3-SLAM.} This is a variant of the 1-line SLAM task 
considered before in which the observation received by the agent when 
in a certain cell  depends also  on    the colors of the two  \emph{adjacent} cells. 
In such version of the problem, sampling the agent location  is no longer sufficient
for decoupling the color cell variables, but in our formulation makes the belief
factors acyclic, so that exact inference over such factors is not exponential in the
total number of variables but in the size of the largest factor. 
The observation is a 0/1 token equal to the color of the current cell with probability 
$p=0.9^{noise}$ where $noise$ is one plus the number of adjacent cells with a different color.
For a line of $n$ cells, there are $n$  beams of size up to 4, one for each cell, that  
include the agent location and the color of the three cells that influence the observation at the
location. The agent location is the only external variable and is not backward
deterministic, so it must be sampled.   Table~\ref{table:1d-slam} shows results 
for instances of size  $n=64$ and $n=512$, using 
16 and  256 particles sampled with the optimal proposal distribution.
The executions choose actions randomly until  each cell of the grid is
visited  10 times. The table shows averages with $97.5\%$ confidence intervals 
over 100 executions for percentages  of cells labeled correctly (one of two colors), 
not labeled at all (no sufficiently certainty), and times per step
and per execution. For this problem, a cell is assumed ``labeled''
when for one of the possible colors its marginal probability is 0.55
of higher.  Since the resulting factorized  model has bounded and small treewidth,
JT scales up well to provide a good baseline with  few errors, although   40\% of the cells are not labeled. 
The quality for BP in this case is similar but runs slower. 
\ACzero\ is an order-of-magnitude faster but of lower quality.
The results for \ACone\ are similar. 

\Omit{
 feasible for any dimension of the problem. BP is also
good but slower than JT, while AC0 and AC1 improve significantly
the performance at the expense of quality in terms of precision and recall.
Statistically, we only differentiate between AC0 and AC1 on instance 1$\times$64
with 256 particles in favor of AC1 with a $p$-level of $2.25\%$.
\Alert{Comparison?}
}

\begin{table}[t]
\centering
\resizebox{\columnwidth}{!}{
\begin{tabular}{@{}cclrrrr@{}}
  \toprule
  1$\times n$ & \#p &   inf. &         \%good &     \%unknowns &   time / step &     time / exec \\
  \midrule
  1$\times$ 64 &  16 & JT     & $97.7\pm0.2$ & $41.7\pm0.7$ & $0.0\pm0.0$ &   $2.9\pm0.0$ \\ 
  1$\times$ 64 &  16 & BP     & $98.5\pm0.2$ & $43.0\pm0.7$ & $0.0\pm0.0$ &   $8.9\pm0.0$ \\ 
  1$\times$ 64 &  16 & \ACzero& $82.9\pm0.6$ & $36.5\pm0.7$ & $0.0\pm0.0$ &   $1.1\pm0.0$ \\ 
  1$\times$ 64 &  16 & \ACone & $84.0\pm0.5$ & $36.5\pm0.7$ & $0.0\pm0.0$ &   $1.1\pm0.0$ \\ 
  \midrule
  1$\times$ 64 & 256 & JT     & $97.9\pm0.2$ & $41.0\pm0.8$ & $0.0\pm0.0$ &  $42.8\pm0.4$ \\ 
  1$\times$ 64 & 256 & BP     & $98.5\pm0.2$ & $40.8\pm0.7$ & $0.1\pm0.0$ & $110.9\pm0.9$ \\ 
  1$\times$ 64 & 256 & \ACzero& $82.5\pm0.6$ & $34.5\pm0.7$ & $0.0\pm0.0$ &  $19.1\pm0.1$ \\ 
  1$\times$ 64 & 256 & \ACone & $83.9\pm0.6$ & $33.3\pm0.8$ & $0.0\pm0.0$ &  $19.6\pm0.1$ \\ 
  \midrule
  1$\times$512 &  16 & JT     & $97.8\pm0.1$ & $47.8\pm0.2$ & $0.0\pm0.0$ & $319.7\pm1.0$ \\ 
  1$\times$512 &  16 & BP     & $98.3\pm0.1$ & $47.8\pm0.2$ & $0.1\pm0.0$ & $681.7\pm2.8$ \\ 
  1$\times$512 &  16 & \ACzero& $81.8\pm0.2$ & $42.3\pm0.2$ & $0.0\pm0.0$ &  $88.7\pm0.3$ \\ 
  1$\times$512 &  16 & \ACone & $81.7\pm0.2$ & $42.1\pm0.2$ & $0.0\pm0.0$ &  $91.0\pm0.3$ \\ 
  \midrule
  1$\times$512 & 256 & JT     & $98.2\pm0.1$ & $47.5\pm0.2$ & $0.7\pm0.0$ & $4,012.9\pm11.5$ \\ 
  1$\times$512 & 256 & BP     & $98.4\pm0.1$ & $47.7\pm0.2$ & $1.4\pm0.0$ & $8,193.1\pm22.6$ \\ 
  1$\times$512 & 256 & \ACzero& $82.7\pm0.2$ & $41.7\pm0.2$ & $0.2\pm0.0$ & $1,117.9\pm3.1$ \\ 
  1$\times$512 & 256 & \ACone & $82.6\pm0.2$ & $42.1\pm0.2$ & $0.2\pm0.0$ & $1,121.0\pm3.0$ \\ 
  \bottomrule
\end{tabular}
}
\caption{PBT in 1-line SLAM using three methods for computing marginals. 
  In this task, observations depend on the color of the current and 
  adjacent cells. Figures are averages  over 100 random executions, each
  of length roughly 10 times the number of cells.
}
\label{table:1d-slam}
\end{table}

\medskip\noindent
\textbf{Minemapping.}
This  problem is a version of Minesweeper that involves an  agent that
moves stochastically and receives noisy information about the presence/absence
of mines in the cell and  surrounding cells. The task for the agent is to map the minefield 
instead of clearing it as in Minesweeper, and the  observations are similar but noisy.
More precisely, movement actions have .9 probability of success and .1 probability of doing nothing,
and when the current location is $i$, the observation token $o$ is generated by summing stochastic
indicator variables $I(j)$  for each adjacent cell $j$ and for $j=i$, where the variable $I(j)$ is
equal to 1 (resp.\ 0) with probability $.9$ if the cell $j$ contains a mine (resp.\ no mine).
The resulting beam structure is similar
to the one for Minesweeper except that the agent location   belongs to all the beams
and must be sampled. Results for this problem are shown Table~\ref{table:dyn-ms}
in a format similar to Table~\ref{table:1d-slam}.  This time, JT is feasible for the small instances only,  
and BP appears to be the best choice for approximating the marginals: it doesn't make many mistakes 
when labeling cells (with mines or not) and runs faster than AC methods that make many more mistakes.

\section{Discussion}

We have introduced a  formulation and algorithm PBT for tracking
probabilistic beliefs in the presence of stochastic actions and sensors
in the form of local belief factors  that can be progressed independently in time
when variables appearing  in more than one  beam  are backward deterministic. 
In such a case, the local belief factors provide an exact decomposition of 
the joint distribution at any time point, and progressing the factors
in time is  exponential in the size of the beams.  The  beams 
are fully determined by the 2-DBN model structure, and are 
usually bounded and small. For computing marginal probabilities, however,
the local belief factors need to be merged. This computation can be performed exactly  
using the jointree algorithm or approximately  by belief propagation or other local consistency methods. 
In addition, when  the beams share variables that are not backward deterministic, such variables 
must be sampled to make them backward deterministic given their sampled histories.
% We have presented empirical results for PBT using three different methods
% for computing marginals from belief factors. 

\begin{table}[t]
\centering
\resizebox{\columnwidth}{!}{
\begin{tabular}{@{}cclrrrr@{}}
  \toprule
  $n\!\times\!n$ & \#p & inf . &         \%good &     \%unknowns &   time / step  & time /exec \\
  \midrule
    6$\times$6 &  32 & JT      & $97.3\pm0.6$ & $12.1\pm1.6$ & $0.2\pm0.0$ &    $73.6\pm  1.0$ \\ 
    6$\times$6 &  32 & BP      & $98.4\pm0.6$ & $16.6\pm1.8$ & $0.2\pm0.0$ &    $81.0\pm  2.2$ \\ 
    6$\times$6 &  32 & \ACzero & $68.0\pm1.5$ & $18.1\pm0.8$ & $0.1\pm0.0$ &    $57.4\pm  3.2$ \\ 
    6$\times$6 &  32 & \ACone  & $69.0\pm1.4$ & $17.6\pm0.8$ & $0.1\pm0.0$ &    $56.8\pm  3.3$ \\ 
  \midrule
    6$\times$6 & 256 & JT      & $97.9\pm0.6$ &  $8.5\pm1.4$ & $1.7\pm0.0$ &   $515.9\pm  5.9$ \\ 
    6$\times$6 & 256 & BP      & $97.9\pm0.6$ & $10.3\pm1.5$ & $1.6\pm0.0$ &   $506.7\pm  9.8$ \\ 
    6$\times$6 & 256 & \ACzero & $67.0\pm1.6$ & $17.2\pm0.8$ & $1.3\pm0.0$ &   $395.9\pm 20.6$ \\ 
    6$\times$6 & 256 & \ACone  & $66.8\pm1.6$ & $17.4\pm0.7$ & $1.3\pm0.0$ &   $397.8\pm 20.9$ \\ 
  \midrule
  10$\times$10 &  32 & BP      & $98.9\pm0.5$ & $33.9\pm1.3$ & $0.8\pm0.0$ &   $814.6\pm  8.3$ \\ 
  10$\times$10 &  32 & \ACzero & $55.6\pm0.9$ & $23.7\pm0.6$ & $0.9\pm0.0$ &   $852.3\pm 53.3$ \\ 
  10$\times$10 &  32 & \ACone  & $55.0\pm0.9$ & $22.5\pm0.6$ & $0.9\pm0.0$ &   $883.9\pm 48.7$ \\ 
  \midrule
  10$\times$10 & 256 & BP      & $97.4\pm0.6$ & $27.9\pm1.3$ & $5.7\pm0.0$ & $5,326.6\pm 48.9$ \\ 
  10$\times$10 & 256 & \ACzero & $54.7\pm1.1$ & $23.8\pm0.6$ & $7.0\pm0.3$ & $6,531.9\pm333.7$ \\ 
  10$\times$10 & 256 & \ACone  & $54.3\pm1.2$ & $23.5\pm0.7$ & $7.1\pm0.3$ & $6,643.6\pm307.2$ \\ 
  \bottomrule
\end{tabular}
}
\caption{PBT in Minemapping using three methods for computing marginals.  
  Figures are averages over 100 random executions. Each execution consists of
  roughly 5 times the number of cells in the grid.
}
\label{table:dyn-ms}
\end{table}

As far as we know, there are no other general, principled approaches 
for dealing effectively with problems such as  Minemapping or  even 1-line-3-SLAM.
RB particle-filtering methods would need to consider too many particles
for making inference tractable in the first problem, and would need to
be programmed to exploit  the resulting  tractable factorization in the second.
In our setting, this all follows from the problem structure and the general formulation.
At the same time,  decomposition methods that operate over disjoint factors  
result in poor approximations, and  standard grid SLAM algorithms \cite{burgard:slam}
involve a number of domain-dependent tricks that would not apply 
to more general problems, like the idea of associating
one particular map to each particle,  drastically simplifying
the uncertainty about the map. 

In the future, we want to develop new ideas for scaling PBT further
so that it can be applied to more realistic SLAM  problems. 
The bottleneck is not the progression of factors but the computation of 
marginals from factors that is done from scratch at every time point. 
We want to explore ways for making such computation incremental.
Likewise, the performance  of PBT needs to be compared with other
approaches, in particular, approaches that also manage to exploit 
forms of context-specific independence and determinism \cite{belgas:dbns}.

\subsection*{Acknowledgments}
We thank the reviewers for useful comments.
This work is partially funded by grant TIN2015-67959, MEC, Spain.

% The \fsz{} planner is partially built upon the Lightweight Automated Planning Toolkit (\url{www.lapkt.org}) and the Fast Downward \pddl{} parser 
% (\url{www.fast-downward.org}).
% TODO - Acknowledge the contribution of Miquel?

\Omit{ I wrote but removed this ... 
Actually, after using the marginals for computing samples and weights,
the marginals are not used in the progression and thrown away. 
Indeed, it is not correct to replace the factors $B_j^h(x_j|\bar u)$
by the marginals $\P{x_j}{h,\bar u}$ in the computation of the
factors for the next time step. Yet  this does not mean that 
the marginals can't be used for that in combination with 
other marginals. Then if the new observation falls into one
beam, the new marginals could be computed incrementally
from the old marginals, rather than from scratch from the
belief factors. 
}

\Omit{
Power of formulation
Discuss relation to BT and incrementality of AC and BP; 
propagation of projections to speed up BP, 
Relation to Grid SLAM algorithms like Burgard et al.: ad-hoc assumptions
  about scan matching, integrate scan, etc.
Basic approximation using AC; room for more sophisticated and 
  effective approximations ..
}

\bibliographystyle{named}
\bibliography{control}

\end{document}